\documentclass[11pt,a4paper]{article}
\usepackage[colorlinks=true,linkcolor=blue]{hyperref}
\usepackage[utf8x]{inputenc}
\usepackage[T1]{fontenc}
\usepackage{lmodern}
\usepackage{amsmath,amsfonts,amssymb,dsfont}
\usepackage[lined,boxed,commentsnumbered]{algorithm2e}
\usepackage{color}
\usepackage{graphicx}
\usepackage{cleveref}
\usepackage{xfrac}
\graphicspath{{./img/}}
\usepackage{microtype}

\renewcommand{\paragraph}[1]{\medskip\noindent\textit{#1}}

\newcommand{\deq}{\mathrel{\mathop:}=}

\newcommand{\E}{\mathbb{E}}
\DeclareMathOperator{\Id}{Id}

\DeclareMathOperator{\diag}{diag}
\newcommand{\transp}[1]{#1^{\!\top}\!}
\def\eps{\varepsilon}
\renewcommand{\epsilon}{\varepsilon}

\title{Natural Langevin Dynamics for Neural Networks}
\author{Gaétan Marceau-Caron\footnote{MILA, Université de Montréal,
Canada} \and Yann Ollivier\footnote{CNRS, TAU, Université Paris-Saclay,
France}}
\date{}

\begin{document}

\maketitle

\begin{abstract}
One way to avoid overfitting in machine learning is to 
use model parameters distributed according to a Bayesian
posterior given the data, rather than the maximum likelihood estimator.
\emph{Stochastic gradient Langevin dynamics} (SGLD) is one 
algorithm to approximate such Bayesian posteriors for large models and
datasets.
SGLD is a standard stochastic
gradient descent to which is added a controlled amount of noise,
specifically scaled so that 
the parameter converges in law to the
posterior distribution \cite{wellingteh2011,teh2016consistency}. The
posterior predictive distribution can be approximated by an ensemble of samples from the
trajectory.

Choice of the variance of the noise is known to impact the practical
behavior of SGLD: for instance, noise should be smaller for sensitive
parameter directions. Theoretically, it has been suggested to use the
inverse Fisher information matrix of the model as the variance of the
noise, since it is also the variance of the Bayesian posterior
\cite{patterson2013langevinsimplex,ahnkorattikarawelling2012,girolamicalderhead2011}.
But the
Fisher matrix is costly to compute for large-dimensional models.

Here we use the easily computed Fisher matrix approximations for deep
neural networks from \cite{marceau2016,Ollivier2015}. The
resulting \emph{natural Langevin dynamics} combines the advantages of
Amari's natural gradient descent and Fisher-preconditioned Langevin
dynamics for large neural networks.

Small-scale experiments on MNIST show that Fisher matrix
preconditioning brings SGLD  close to dropout as a regularizing
technique.
\end{abstract}

\newcommand{\deltat}{\ensuremath{\hspace{0.05em}\delta\hspace{-.06em}t\hspace{0.05em}}}
\newcommand{\gaussian}{\mathcal{N}}
\newcommand{\data}{\mathcal{D}}
\newcommand{\thetamean}{\bar \theta}




Consider a supervised learning problem with a
dataset
$\data=\{(x_1,y_1),\ldots,\allowbreak (x_N,y_N) \}$ of $N$ input-output pairs, to be modelled by a parametric
probabilistic distribution $y_i\sim p_\theta(y|x_i)$ ($x=\varnothing$
amounts to unsupervised learning of $y$). 
 Defining the
log-loss
$\ell_\theta(y_i|x_i)\deq -\ln p_\theta(y_i|x_i)$,
the maximum likelihood estimator is the value $\theta$ that minimizes
$\E_{(x,y)\in \data} \ell_\theta(y|x)$, where $\E_{(x,y)\in \data}$ denotes averaging
over the dataset.

Stochastic gradient descent is often used to
tackle this minimization problem for large-scale datasets
\cite{bottoulecun2003,bottou2010sgd}. This consists in iterating
\begin{equation}
\theta\gets \theta -\eta\, \hat\E_{(x,y)\in \data}\,\partial_\theta
\ell_\theta(y|x),
\end{equation}
where $\eta$ is a step size,
$\partial_\theta$ denotes the gradient of a function with respect
to $\theta$, and
$\hat\E_{(x,y)\in \data}$ denotes an empirical average of gradients
from a random subset of the dataset $\data$ (a minibatch, which may be of
size $1$). 

Estimating the model parameter $\theta$ via
maximum likelihood, i.e., minimizing the training loss on $\data$,
is prone to overfitting.
Bayesian methods arguably
offer a protection against overfitting (\cite[3.4]{Bishop_book},
\cite[44.4]{Mackay_book};
see also 
\cite{neal1996bayesianNN,mackay1992practical}
for Bayesian neural networks). 
Arguably, the
variance of the posterior distribution of $\theta$ represents the
intrinsic uncertainty on $\theta$ given the
data, and optimizing $\theta$ beyond that point results in overfitting
\cite{wellingteh2011};
sampling the parameter $\theta$ from its Bayesian posterior prevents using
a
too precisely tuned value.

\emph{Stochastic gradient Langevin dynamics} (SGLD)
\cite{wellingteh2011,teh2016consistency} modifies sto\-cha\-stic
gradient descent to provide random values of $\theta$ that
are distributed according to a Bayesian posterior. This is achieved by adding
controlled noise to the gradient descent,
together with an $O(1/N)$ pull 
towards a Bayesian prior:
\begin{equation}
\label{eq:SGLD}
\theta\gets \theta -\eta \,\hat\E_{(x,y)\in \data}\,\partial_\theta
\left(\ell_\theta(y|x)
-\frac{1}{N}\ln
\alpha(\theta)\right)+\sqrt{\frac{2\eta}{N}}\,\gaussian(0,\Id)
\end{equation}
where $N$ is the size of the dataset,
$\alpha(\theta)$ is the density of a Bayesian prior on $\theta$,
and $\gaussian(0,\Id)$ is a random
Gaussian vector of size $\dim(\theta)$. \footnote{
Our convention for the step size $\eta$ differs from
\cite{teh2016consistency} by a factor $2/N$, namely,
$\delta=\frac{2}{N}\eta$ where $\delta$ is the step size in
\cite[(3)]{teh2016consistency}: this allows for a direct
comparison with stochastic gradient descent.
} The
larger $N$ is, the closer SGLD is to simple stochastic gradient descent,
as the Bayesian posterior concentrates around a single point.
The Bayesian
interpretation determines the necessary amount of noise
depending on step size and dataset size. SGLD has the same algorithmic complexity as simple stochastic gradient
descent.

Thanks to the injected noise, $\theta$ does not
converge to a single value, but its \emph{distribution} at time $t$
converges to the Bayesian posterior of $\theta$ given the data, namely,
$\pi(\theta)\propto \alpha(\theta)\prod_{(x,y)\in \data} p_\theta(y|x)$.
A formal proof is given in
\cite{teh2016consistency,chen2015convergence} for suitably decreasing
step sizes; the asymptotically
optimal step size
is $\eta_k\approx k^{-1/3}$ at step $k$, thus, larger than
the usual Robbins--Monro criterion for stochastic gradient descent. The
asymptotic behavior is well understood from
\cite{teh2016consistency,chen2015convergence}, and
\cite{MDM17langevinconv,DurmusMoulines2016langevin} provide sharp
non-asymptotic rates
in the convex case.


One can then extract information from the distribution of
$\theta$. For instance, the Bayesian posterior mean can be
approximated by averaging $\theta$ over the trajectory.
The full Bayesian posterior prediction can be approximated by ensembling
\cite[7.12]{GBCdeeplearning}
predictions from several values of $\theta$ sampled from the
trajectory, though this creates additional computational and memory costs at test time.

We refer to \cite{wellingteh2011,teh2016consistency} for a general discussion of SGLD
(and other Bayesian methods) for large-scale machine learning.


\paragraph{Practical remarks.}
%
For regression problems, the square loss 
$(y-\hat
y(\theta))^2$
between observations $y$ and predictions $\hat y(\theta)$ must be
properly cast as the log-loss of a Gaussian model, $\ell=(y-\hat
y(\theta))^2/2\sigma^2+\dim(y)\ln \sigma$ for a proper choice of
$\sigma$ (such as the empirical RMSE). Just using $\sigma^2=1$ amounts to
using a badly specified error model and will provide a poor Bayesian posterior.
%

The variance coming from computing gradients on a minibatch from
$\data$, $\hat \E_{(x,y)\in\data} \partial_\theta \ell_\theta(y|x)$, adds up to
the SGLD noise. For small step sizes,
$\eta\ll \sqrt{\eta}$, so the SGLD noise dominates.
\cite{ahnkorattikarawelling2012} suggest a
correction for large $\eta$. 

A popular choice of prior $\alpha(\theta)$ is a Gaussian prior
$\gaussian(0,\Sigma^2)$; the variance
$\Sigma^2$ becomes an additional hyperparameter. In line with Bayesian 
philosophy we also tested the conjugate prior for Gaussian distributions
with unknown variance (a
mixture of Gaussian priors for all $\Sigma^2$), the normal-inverse gamma,
with default hyperparameters; empirically,
performance comes close enough to the best $\Sigma^2$, without having to optimize over  
$\Sigma^2$.

\paragraph{Preconditioning the noise.}
SGLD as above introduces uniform noise in all
parameter directions. This might hurt the optimization process. If
performance is more sensitive in certain parameter
directions,
adapting the noise covariance can largely improve SGLD performance.
This
requires changing both the noise covariance \emph{and} the gradient step
by the same matrix
\cite{wellingteh2011,girolamicalderhead2011,ahnkorattikarawelling2012,Li:2016}.

For any positive-definite symmetric matrix $C$, the \emph{preconditioned
SGLD},
\begin{align}
\label{eq:CSGLD}
\theta&\gets 
\theta -\eta \,C\,\hat\E_{(x,y)\in \data}\,\partial_\theta
\left(\ell_\theta(y|x)
-\frac{1}{N}\ln
\alpha(\theta)\right)+\sqrt{\frac{2\eta}{N}}\,C^{1/2} \gaussian(0,\Id)
\end{align}
still converges in law to the Bayesian posterior (it is equivalent to a
non-precon\-ditioned Langevin dynamics on $C^{-1/2}\theta$).
A diagonal $C$ amounts to having distinct
values of the step size $\eta$ for each parameter direction, both for
noise and gradient.

This assumes that $C$ is fixed and does not depend on $\theta$. 
\footnote{
If $C(\theta)$ depends on $\theta$, 
the algorithm involves
derivatives of $C(\theta)$
with respect to $\theta$
\cite{girolamicalderhead2011,xifara2014langevin}. In our case (neural
networks), these are
not readily available.
}
In practice, this means $C$ should be adapted slowly in the algorithms
(hence our use of running averages for $C$ hereafter); the resulting bias
is analyzed in
\cite[Cor.~2]{Li:2016}.

\cite{Li:2016} apply preconditioned SGLD to neural networks, with a
diagonal preconditioner $C$ taken from the RMSProp optimization scheme, a
classical tool to adapt step sizes for each direction of $\theta$.
\footnote{\label{ft:bug}We could not reproduce the good results from \cite{Li:2016}. Their
code contains a bug which produces noise of variance $2\eta/N^2$
instead of $2\eta/N$ in \eqref{eq:SGLD}, thus greatly suppressing
the Langevin noise, and not matching the Bayesian
posterior.}

\paragraph{Langevin preconditioners and information geometry.}
In order to provide a good or even optimal
preconditioner $C$, it has been suggested to set $C$ to
the inverse of the Fisher information matrix
\cite{girolamicalderhead2011,ahnkorattikarawelling2012,patterson2013langevinsimplex}.

The Fisher information matrix $J(\theta)$ at $\theta$, for a model
$p_\theta$, is defined by
\begin{equation}
\label{eq:fish}
J(\theta)\deq \E_{(x,y)\in\data}\,\E_{\tilde y\sim p_\theta(\tilde y|x)} \left[\left(\partial_\theta \ln
		p_\theta(\tilde y|x)\right)\transp{\left(\partial_\theta \ln
		p_\theta(\tilde y|x)\right)}\,\right]
\end{equation}
(note that for supervised learning, we fix the distribution of the inputs $x$ from
the data but sample $y$ according to the model $p_\theta(y|x)$).
Intuitively, the entries of the Fisher
matrix represent the sensitivity of the model in
each parameter direction.

Using the inverse Fisher matrix as the SGLD preconditioner $C$ has
several theoretical advantages.
First, this
reduces Langevin noise in sensitive parameter
directions (thanks to the Fisher matrix being the average of squared
gradients).

Second, since $C$ also affects the gradient term in \eqref{eq:CSGLD}, the
gradient part of SGLD becomes Amari's \emph{natural gradient}, 
known to have theoretically optimal convergence \cite{Amari1998}. The
resulting algorithm is also insensitive to changes of variables in
$\theta$ (for small learning rates) and makes sense if
$\theta$ belongs to a manifold.

\newcommand{\thetamap}{\theta^{\ast}}

Third, the Bayesian posterior variance of the parameter $\theta$ is
asymptotically proportional to the inverse Fisher information matrix
$J(\thetamap)^{-1}$ at the maximum a posteriori $\thetamap$
(Bernstein--von Mises theorem \cite{van2000asymptotic}).
So with Fisher preconditioning, the noise injected in the
optimization process has the same shape as the actual noise in the target
distribution on $\theta$.
Thus, it is tempting to investigate the behavior of SGLD with noise
covariance
$C\propto
J(\thetamap)^{-1}$. 


\paragraph{Approximating the Fisher matrix for large models.}
The Fisher matrix $J(\thetamap)$ can be estimated by replacing the
expectation in its definition \eqref{eq:fish} by an empirical
average along the trajectory
\cite{ahnkorattikarawelling2012}. This results in Algorithm~\ref{alg:fullfisher}
below.\footnote{The Fisher matrix definition \eqref{eq:fish} averages
over synthetic data $\tilde y$ generated by
$p_\theta(\tilde y|x)$.
In practice, using the samples $y$ from the
dataset is simpler (the OP variant in Alg.~\ref{alg:fullfisher}). This can result in significant differences
\cite{marceau2016,Ollivier2015,pascanu:2013}, even in simple cases.}

However, for large-dimensional models such as deep neural networks, the
Fisher matrix is too large to be inverted or even stored (it is a full
matrix of size $\dim(\theta)\times \dim(\theta)$). So
approximation strategies are necessary.

Approximating the Fisher matrix does not invalidate asymptotic
convergence of SGLD, since \eqref{eq:CSGLD} converges to the true Bayesian
posterior for any preconditioning matrix $C$. But 
the closer $C$ is to the inverse
Fisher matrix, the closer SGLD will be to a natural gradient descent, and
SGLD noise to the true posterior variance.


One way of building principled approximations of the Fisher matrix is to
reason in terms of the associated invariance group. The full Fisher
matrix provides invariance
under all changes of variables in parameter space $\theta$:
optimizing by natural gradient descent over $\theta$ or over a
reparameterization of $\theta$ will yield the same learning trajectories
(in the limit of small learning rates). Meanwhile, the Euclidean
gradient descent does not have any invariance properties (e.g., 
inverting black and white in the image inputs of a neural network affects
performance). We refer to \cite{Ollivier2015} for further discussion in
the context of neural networks.

The diagonal of the Fisher matrix is the most
obvious approximation. Its invariance subgroup consists of all rescalings of
individual parameter components.

The \emph{quasi-diagonal} approximation of the Fisher matrix
\cite{Ollivier2015} is built to retain more invariance properties of the
Fisher matrix, at a small computational cost. It
provides
invariance under all affine transformations of the \emph{activities} of
units in a neural network (e.g., shifting or rescaling the inputs, or
switching from sigmoid to tanh activation function). 
The quasi-diagonal approximation maintains the diagonal of the Fisher
matrix plus a few well-chosen off-diagonal terms,
requiring to
store an additional vector of size $\dim(\theta)$. Overall, the resulting
algorithmic complexity is of the same order as ordinary backpropagation,
thus suitable for large-dimensional models.
\cite{Ollivier2015} also
provides more complex approximations with a larger invariance group,
suited to sparsely connected neural networks.

The resulting \emph{quasi-diagonal natural gradient} can be coded efficiently
\cite{marceau2016}; experimentally,
the few extra off-diagonal terms can make a
large difference.

%
%

\paragraph{Natural Langevin dynamics for neural networks:
implementation.}
Algorithm~\ref{alg:langevinC}
presents the Langevin dynamics with a generic preconditioner $C$. For the
ordinary SGLD, $C$ would be the identity
matrix. The
internal setup of a preconditioner decouples from the general
implementation of SGLD optimization.
A preconditioner $C$ is a matrix object that provides
the routines needed by Algorithm~\ref{alg:langevinC}:
\\
\indent -- Multiply a gradient estimate by $C$: $g\gets Cg$;\\
\indent -- Draw a Gaussian random vector 
$\xi\sim \gaussian(0,C)=C^{1/2}\gaussian(0,\Id)$;\\
\indent -- Update $C$ given recent gradient observations;\\
\indent -- An initialization procedure for $C$ at startup.
\\
We
now make these routines explicit for several choices of
preconditioner.

The RMSProp preconditioner used in \cite{Li:2016}
divides gradients by their recent magnitude: $C$ is
diagonal, and for each
parameter component $i$, $C_{ii}$ is the inverse of a root-mean-square
average of recent gradients in direction $i$ (Alg.~\ref{alg:rmsprop}).

Algorithm~\ref{alg:fullfisher} describes preconditioned SGLD with a
preconditioner $C=J^{-1}$ using the full Fisher matrix $J$ at the
posterior mean $\theta^\ast$. This is suitable only for small-dimensional
models. The Fisher matrix is obtained as a moving average of rank-one
contributions over the trajectory
(Alg.~\ref{alg:fullfisher}). This moving average has the
further advantage of smoothing the fluctuations of the parameter
$\theta$ over the SGLD trajectory, ensuring convergence \cite{ahnkorattikarawelling2012}.  


Finally we consider SGLD using the \emph{quasi-diagonal} Fisher matrix, the object
of the tests in this article, applicable to large-dimensional
models.

For a neural network, the parameters are
grouped into blocks corresponding to the bias and incoming weights of each
neuron, with the bias being the first parameter in a block. The
Fisher matrix $J$ is updated as in Algorithm~\ref{alg:fullfisher}, but
storing only its diagonal and the first row in each block. Then a
Cholesky decomposition $C=A\transp{A}$ is maintained for the
preconditioner $C$, such that
the axioms of the
quasi-diagonal approximation are satisfied (Algorithm \ref{alg:qdchol}): in each block, $A$
has non-zero entries only on its diagonal and first row, and is built such
that $C^{-1}=(\transp{A})^{-1}A^{-1}$ has the same first row and diagonal
as the Fisher matrix $J$.
The sparse Cholesky decomposition provides the operations
of the 
preconditioner: multiplying by $C=A\transp{A}$ and sampling from
$\gaussian(0,C)=A\,\gaussian(0,\Id)$.

\paragraph{Experiments.}
We compare empirically four SGLD preconditioners: Euclidean ($C=\Id$,
standard SGLD),
RMSProp, Diagonal Outer Product (DOP) and Quasi-Diagonal Outer Product
(QDOP) on the MNIST dataset.
The Euclidean and RMSProp results widely mismatch those from 
\cite{Li:2016}, see footnote \ref{ft:bug}.

We compare SGLD to Dropout, a standard regularization procedure for
neural networks. For SGLD we compare the performance of using a single
network set to the posterior mean, and an ensemble of networks sampled
from the trajectory (theoretically closer to the true Bayesian posterior, but
computationally costlier).

The code for the experiments can be found at\\
{\footnotesize\url{https://github.com/gmarceaucaron/natural-langevin-dynamics-for-neural-networks}} .
We use a feedforward ReLU network with two hidden layers of size $400$,
with the usual $\gaussian(0,1/\text{fan-in})$ initialization
\cite{GBCdeeplearning}. Inputs are normalized to $[0;1]$. Step sizes
are optimized over $\eta \in \{ .001, .01, .1,1\}$
 for Euclidean and $\eta \in \{ .0001,.001,.01,.1\}$
for the others,
with schedule $\eta\gets \eta/2$ every 10,000 updates \cite{Li:2016}.
Minibatch size is $100$. The metric decay rate and regularizer are
$\gamma_t=1/\sqrt{t}$ and $\eps=10^{-4}$. The prior was a
Gaussian $\gaussian(0,\sigma^2)$ with $\sigma^2 \in \{0.01, 0.1, 1\}$.
The Bayesian posterior ensemble is built by storing every $100$-th
parameter value of the trajectory after the first $500$.



\begin{table}
  \centering
  \begin{tabular}{lcccc}
    Method & NLL (train) & Accuracy (train) & NLL (test) & Accuracy (test)\\
    \hline 
    SGD & 0.0003 &     100.00 &      0.0584 &       98.24 \\
    Dropout  & 0.0006 &     100.00 &      0.0519 &       98.61 \\
    \hline \hline
    Ensemble, Euclidean & 0.0357 &      99.63 &     0.0726 &      98.10 \\
    Ensemble, RMSProp & 0.0415 &      99.47 &     0.0742 &      98.17 \\
    Ensemble, DOP & 0.0292 &      99.69 &     0.0660 &      98.13 \\
    Ensemble, QDOP & 0.0229 &      99.85 &     0.0591 &      98.38 \\
    \hline
    PostMean, Euclidean & 0.0281 &      99.12 &     0.1240 &      97.16 \\
    PostMean, RMSProp &  0.0299 &      99.07 &     0.1134 &      97.21 \\
    PostMean, DOP & 0.0243 &      99.20 &     0.1389 &      97.20 \\
    PostMean, QDOP & 0.0292 &      99.60 &     0.3429 &      98.14 \\
    \hline
  \end{tabular}
  \caption{\label{tb:results}Performance on the MNIST test set with a feedforward 400-400
    architecture. \label{tab:results} Hyperparameters were selected based
    on accuracy on a validation set. The methods are SGD without regularization,
  Dropout, SGLD ensemble and SGLD posterior mean (PostMean) with a
  Gaussian prior ($\sigma^2=0.1$).}
\end{table}

Table \ref{tb:results} shows that SGLD with a
quasi-diagonal Fisher matrix preconditioner and Bayesian posterior
ensembling outperforms other SGLD settings.

Bayesian theory favors the use of the full Bayesian posterior at test
time, rather than any single parameter value. The results here are
consistent with this viewpoint: 
using a single parameter set to the Bayesian posterior mean offers much
poorer performance than either Dropout or a Bayesian posterior
ensemble. (Dropout also has a Bayesian inspiration as a mixture of
models \cite{srivastava2014dropout}.) This is also consistent with the
generally good performance of ensemble methods.

All other preconditioners perform worse than QDOP or Dropout. In
particular, the diagonal Fisher matrix offers no advantage over RMSProp,
while the
\emph{quasi-diagonal} Fisher matrix does. This is consistent with
\cite{marceau2016} and may vindicate the quasi-diagonal construction
via an invariance group viewpoint.

\begin{algorithm}
  \SetAlgoLined
  \KwData{Dataset $\data=\{(x_1,y_1),\ldots,(x_N,y_N)\}$ of size $N$\;
  probabilistic model $p_\theta(y|x)$ with log-loss $\ell(y|x)\deq -\ln
  p_\theta(y|x)$\;
  Bayesian prior $\alpha(\theta)=\gaussian(\theta_0,\Sigma_0)$, default:
  $\theta_0=0$\;
  Learning rate $\eta_t\ll 1$. Preconditioner $C$ (for simple SGLD: $C=\Id$).
  }
  \KwResult{Parameter $\theta$ whose distribution approximates the
  Bayesian posterior $\Pr(\theta \mid D,\alpha)$. Approximation
  $\thetamean$ of the Bayesian posterior mean of $\theta$.}
  \SetKwFunction{qdrankoneupdate}{QDRankOneUpdate}
  \SetKwFunction{qdsolve}{QDSolve}
  \SetKw{init}{Initialization:}
  \init{} $\theta\sim \alpha(\theta)$; $\thetamean\gets \theta_0$; initialize preconditioner\;
  \While{not finished}{
      retrieve a data sample $x$ and corresponding target $y$ from $\data$\;
      forward $x$ through the network, and compute loss $\ell(y|x)$\;
      backpropagate and compute gradient of loss: $g\gets
      \partial_\theta \ell(y|x)$ (for a minibatch: let $g$ be the
      \emph{average}, not the sum, of individual gradients)\;
      incorporate gradient of prior: $g\gets
      g+\frac{1}{N}\Sigma_0^{-1}(\theta-\theta_0)$\;
      update preconditioner $C$ using current sample and gradient $g$\;
      apply preconditioner: $g\gets Cg$\;
      sample preconditioned noise: $\xi \sim
      \gaussian(0,C)=C^{1/2}\gaussian(0,\Id)$\;
      update parameters: $\theta\gets \theta-\eta_t\,g
      +\sqrt{(2\eta_t/N)}\, \xi$\;
      update posterior mean: $\thetamean\gets
      (1-\mu_t)\thetamean+\mu_t\theta$.
  }
  \caption{SGLD with a generic
  preconditioner $C$. For instance $C$ may be $\Id$ (Euclidean SGLD), a
  diagonal preconditioner such as RMSProp, the inverse of a
  Fisher matrix approximation...\label{alg:langevinC}}
\end{algorithm}

\begin{algorithm}
  \SetAlgoLined
  \KwData{Preconditioner $C=D^{-1/2}$ with $D$ a diagonal matrix of size
  $\dim(\theta)$; decay rate $\gamma_t$; regularizer $\eps\geq 0$.}
  \SetKw{init}{Initialization:}
  \init{} {$D\gets \diag(1)$\;}
  \SetKw{Cupdate}{Preconditioner update:}
  \Cupdate{} {$D_{ii}\gets (1-\gamma_t)D_{ii}+\gamma_t \,g_i^2$ with $g_i$
  the components of the gradient
  of the current sample\;}
  \SetKw{Capply}{Preconditioner application:}
  \Capply{} $g_i \gets (D_{ii}+\eps)^{-1/2}\,g_i$\;
  \SetKw{Csample}{Preconditioned noise:}
  \Csample{} $\xi_i \gets
  (D_{ii}+\epsilon)^{-1/4}\gaussian(0,1)$.
  \caption{RMSProp routines for SGLD, similar to \cite{Li:2016}.
  \label{alg:rmsprop}}
\end{algorithm}

\begin{algorithm}
  \SetAlgoLined
  \KwData{Preconditioner $C=J^{-1}$ with $J$ the Fisher matrix; decay
  rate $\gamma_t$; regularizer $\eps\geq 0$.}
  \SetKw{init}{Initialization:}
  \init{} {$J\gets \diag(1)$\;} 
  \SetKw{Cupdate}{Preconditioner update:}
  \Cupdate{} {Synthesize output $\tilde y\sim p_\theta(\tilde y|x)$ given
  current model $\theta$ and current input $x$ (OP variant: just
  use $\tilde y=y$ from the dataset)\;
  Compute gradient of loss for $\tilde y$: $\tilde v\gets \partial_\theta
  \ell(\tilde y|x)$\;
  Update Fisher matrix: $J\gets (1-\gamma_t)J+\gamma_t \tilde v
  \transp{\tilde v}$\;} 
  \SetKw{Capply}{Preconditioner application:}
  \Capply{} $v \gets (J+\eps \Id)^{-1} v$\;
  \SetKw{Csample}{Preconditioned noise:}
  \Csample{} $\xi \gets
  (J+\eps \Id)^{-1/2}\gaussian(0,\Id)$.
  \caption{Routines for SGLD with full Fisher matrix.
  \label{alg:fullfisher}}
\end{algorithm}

\begin{algorithm}
    \SetAlgoLined
    \KwData{Symmetric positive matrix $J$ of which only the diagonal and
    first row are known; regularizer $\eps \geq 0$.}
    \KwResult{Sparse matrix $A$ whose non-zero entries lie only on the diagonal
    and first row, and such that $(\transp{A})^{-1}A^{-1}$ has the same
    diagonal and first row as $J+\eps\Id$.}
    $A\gets 0$; $A_{00}\gets \frac{1}{\sqrt{J_{00}+\eps}}$
    (Matrix indices start at $0$)\;
      $A_{ii}\gets \frac{1}{\sqrt{J_{ii} - (A_{00} J_{0i})^2 + \eps}}$
      for each index $i\neq 0$\;
      $A_{0i}\gets -A_{00}^2 A_{ii} J_{0i}$ for each index $i\neq 0$\;
    \Return $A$\;
    \caption{Quasi-diagonal Cholesky decomposition.  \label{alg:qdchol}}
  \end{algorithm}

\bibliographystyle{alpha}
\bibliography{library}

\newcommand{\etalchar}[1]{$^{#1}$}
\begin{thebibliography}{SHK{\etalchar{+}}14}

\bibitem[AKW12]{ahnkorattikarawelling2012}
Sungjin Ahn, Anoop Korattikara, and Max Welling.
\newblock Bayesian posterior sampling via stochastic gradient {F}isher scoring.
\newblock In {\em ICML}, 2012.

\bibitem[Ama98]{Amari1998}
Shun-ichi Amari.
\newblock Natural gradient works efficiently in learning.
\newblock {\em Neural Comput.}, 10:251--276, February 1998.

\bibitem[Bis06]{Bishop_book}
C.~M. Bishop.
\newblock {\em Pattern recognition and machine learning}.
\newblock Springer, 2006.

\bibitem[BL03]{bottoulecun2003}
L{\'e}on Bottou and Yann LeCun.
\newblock Large scale online learning.
\newblock In {\em NIPS}, volume~30, page~77, 2003.

\bibitem[Bot10]{bottou2010sgd}
L{\'e}on Bottou.
\newblock Large-scale machine learning with stochastic gradient descent.
\newblock In {\em Proceedings of COMPSTAT'2010}, pages 177--186. Springer,
  2010.

\bibitem[CDC15]{chen2015convergence}
Changyou Chen, Nan Ding, and Lawrence Carin.
\newblock On the convergence of stochastic gradient {MCMC} algorithms with
  high-order integrators.
\newblock In {\em Advances in Neural Information Processing Systems}, pages
  2278--2286, 2015.

\bibitem[DM16]{DurmusMoulines2016langevin}
Alain Durmus and Eric Moulines.
\newblock High-dimensional {B}ayesian inference via the unadjusted {L}angevin
  algorithm.
\newblock arXiv preprint arXiv:1605.01559, 2016.

\bibitem[GBC16]{GBCdeeplearning}
Ian Goodfellow, Yoshua Bengio, and Aaron Courville.
\newblock {\em Deep learning}.
\newblock MIT press, 2016.

\bibitem[GC11]{girolamicalderhead2011}
Mark Girolami and Ben Calderhead.
\newblock Riemann manifold {L}angevin and {H}amiltonian {M}onte {C}arlo
  methods.
\newblock {\em Journal of the Royal Statistical Society: Series B (Statistical
  Methodology)}, 73(2):123--214, 2011.

\bibitem[LCCC16]{Li:2016}
Chunyuan Li, Changyou Chen, David~E. Carlson, and Lawrence Carin.
\newblock Preconditioned stochastic gradient {L}angevin dynamics for deep
  neural networks.
\newblock In Dale Schuurmans and Michael~P. Wellman, editors, {\em Proceedings
  of the Thirtieth {AAAI} Conference on Artificial Intelligence, February
  12-17, 2016, Phoenix, Arizona, {USA.}}, pages 1788--1794. {AAAI} Press, 2016.

\bibitem[Mac92]{mackay1992practical}
David~JC MacKay.
\newblock A practical {B}ayesian framework for backpropagation networks.
\newblock {\em Neural computation}, 4(3):448--472, 1992.

\bibitem[Mac03]{Mackay_book}
David~JC MacKay.
\newblock {\em Information theory, inference and learning algorithms}.
\newblock Cambridge university press, 2003.

\bibitem[MDM17]{MDM17langevinconv}
Szymon Majewski, Alain Durmus, and Błażej Miasojedow.
\newblock 2017.

\bibitem[MO16]{marceau2016}
Ga{\'{e}}tan Marceau{-}Caron and Yann Ollivier.
\newblock Practical {R}iemannian neural networks.
\newblock {\em arXiv}, abs/1602.08007, 2016.

\bibitem[Nea96]{neal1996bayesianNN}
Radford~M. Neal.
\newblock {\em Bayesian learning for neural networks}.
\newblock Springer, 1996.

\bibitem[Oll15]{Ollivier2015}
Yann Ollivier.
\newblock Riemannian metrics for neural networks {I}: feedforward networks.
\newblock {\em Information and Inference}, 4(2):108--153, 2015.

\bibitem[PB13]{pascanu:2013}
Razvan Pascanu and Yoshua Bengio.
\newblock Natural gradient revisited.
\newblock {\em arXiv}, abs/1301.3584, 2013.

\bibitem[PT13]{patterson2013langevinsimplex}
Sam Patterson and Yee~Whye Teh.
\newblock Stochastic gradient {R}iemannian {L}angevin dynamics on the
  probability simplex.
\newblock In {\em Advances in Neural Information Processing Systems}, pages
  3102--3110, 2013.

\bibitem[SHK{\etalchar{+}}14]{srivastava2014dropout}
Nitish Srivastava, Geoffrey~E Hinton, Alex Krizhevsky, Ilya Sutskever, and
  Ruslan Salakhutdinov.
\newblock Dropout: a simple way to prevent neural networks from overfitting.
\newblock {\em Journal of Machine Learning Research}, 15(1):1929--1958, 2014.

\bibitem[TTV16]{teh2016consistency}
Yee~Whye Teh, Alexandre~H Thiery, and Sebastian~J Vollmer.
\newblock Consistency and fluctuations for stochastic gradient {L}angevin
  dynamics.
\newblock {\em Journal of Machine Learning Research}, 17(7):1--33, 2016.

\bibitem[vdV00]{van2000asymptotic}
A.W. van~der Vaart.
\newblock {\em Asymptotic statistics}.
\newblock Cambridge university press, 2000.

\bibitem[WT11]{wellingteh2011}
Max Welling and Yee~Whye Teh.
\newblock Bayesian learning via stochastic gradient {L}angevin dynamics.
\newblock In {\em Proceedings of the 28th International Conference on Machine
  Learning (ICML-11)}, pages 681--688, 2011.

\bibitem[XSL{\etalchar{+}}14]{xifara2014langevin}
Tatiana Xifara, Chris Sherlock, Samuel Livingstone, Simon Byrne, and Mark
  Girolami.
\newblock Langevin diffusions and the {M}etropolis-adjusted {L}angevin
  algorithm.
\newblock {\em Statistics \& Probability Letters}, 91:14--19, 2014.

\end{thebibliography}

\end{document}